\documentclass{article}

 \usepackage[preprint]{neurips_2026}

\usepackage[utf8]{inputenc} 
\usepackage[T1]{fontenc}    
\usepackage{hyperref}       
\usepackage{url}            
\usepackage{booktabs}       
\usepackage{amsfonts}       
\usepackage{nicefrac}       
\usepackage{microtype}      
\usepackage{xcolor}         

\usepackage{amsmath} 
\usepackage{subcaption}
\usepackage{graphicx}
\usepackage{multirow}
\usepackage{float}
\usepackage{makecell}
\usepackage{colortbl}
\usepackage{xcolor}
\usepackage{booktabs}
\usepackage{array}
\usepackage{wrapfig}
\usepackage{caption}
\usepackage{titletoc}
\definecolor{Cerulean}{RGB}{42,82,190}

\newcommand{\ourMethod}{Metis}

\title{{\ourMethod}: A Generalizable and Efficient World-Action Model for Autonomous Driving and  Urban Navigation}

\author{
Jingyu Li$^{1,2}$\thanks{Equal contribution}\quad
Zhe Liu$^{3}$\footnotemark[1]\quad
Dongnan Hu$^{4,2}$\quad
Junjie Wu$^{5}$\quad
Zipei Ma$^{1,2}$\\
\bfseries
Wenxiao Wu$^{6,2}$\quad
Chao Han$^{5}$\quad
Zhihui Hao$^{5}$\quad
Zhikang Liu$^{5}$\quad
Kun Zhan$^{5}$,\\
\bfseries
Jiankang Deng$^{7}$\quad
Xiatian Zhu$^{8}$\quad
Li Zhang$^{1,2}$\thanks{Corresponding author.}
\\
$^{1}$Fudan University \quad
$^{2}$Shanghai Innovation Institute \quad
$^{3}$The University of Hong Kong \\
$^{4}$Tongji University \quad
$^{5}$Li Auto Inc. \quad
$^{6}$Huazhong University of Science and Technology \\
$^{7}$Imperial College London \quad
$^{8}$University of Surrey
\\[0.5ex]
{
    \href{https://github.com/LogosRoboticsGroup/Metis}
{\textcolor{Cerulean}{\texttt{github.com/LogosRoboticsGroup/Metis}}}
    }
}

\begin{document}

\maketitle

\begin{abstract}
 
World action models~(WAMs) have shown great promise for autonomous driving and urban navigation.
Built upon Vision-Language-Action models or video generation models,
existing approaches suffer key limitations: (1) High inference latency due to future observation prediction at test time, and (2) tightly coupled video and action modeling leading to representational mismatch and degraded generalization.
To address both issues, we propose \textbf{\ourMethod}, an end-to-end WAM framework that decouples video generation and action prediction. 
Specifically, {\ourMethod} employs a Mixture-of-Transformers architecture with dedicated experts for video generation and action prediction, preserving the intrinsic distributional properties of each task.
To enhance efficiency, we introduce an asymmetric attention mask that enables joint training of both experts while allowing the action model to bypass explicit video generation during inference. This design ensures training-inference consistency and significantly reduces computational costs without compromising planning performance.
Extensive experiments demonstrate state-of-the-art performance on the NAVSIM navhard and navtest benchmarks and the CityWalker navigation benchmark, validating both the generalizability and efficiency across diverse tasks. Real-robot deployments further confirm the practical feasibility of our approach.

\end{abstract}
\section{Introduction}

Achieving safe and rational trajectory planning in Autonomous Driving ~(AD) and Urban Navigation~(UN) requires policies that can not only react to current observations, but also anticipate how the environment evolves under agent interactions. This perspective has led to the emergence of World action models~(WAMs), which integrate action generation with predictive modeling of future observations within a unified framework~\cite{drive-WM,drivelaw,epona,pwm}. By modeling future outcomes, WAMs provide a natural way to capture physical dynamics and task-relevant temporal dependencies, offering a more expressive alternative to standard Vision-Language-Action~(VLA) models~\cite{recogdrive,EMMA,AutoVLA,omnivla}.

\begin{figure}[t]
    \centering
    \includegraphics[width=1\linewidth]{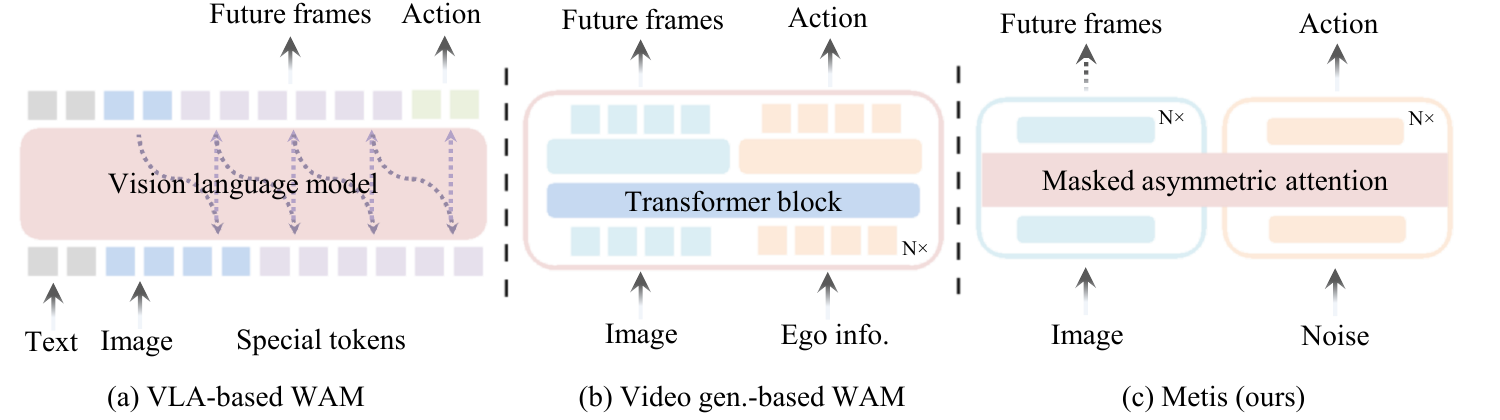}
    \caption{Different WAM paradigms. (a)~VLA-based WAMs rely on autoregressive token-based future prediction for action planning. (b)~Video generation-based WAMs use tightly coupled architectures to jointly predict future video and actions. (c)~Our \textbf{\ourMethod} decouples video generation from action inference via an masked asymmetric attention, enabling efficient planning without video generation.}
    \label{fig:intro}
    \vspace{-10pt}
\end{figure}

Recent WAMs have been rapidly evolving, giving rise to diverse design paradigms. A natural extension of VLA-based methods toward WAMs is to augment Vision-Language Models~(VLMs)~\cite{gpt4,LLaVA,internvl,qwen25vl} with additional tokens for autoregressive future observation generation, leveraging the rich prior knowledge of VLMs while modeling future environmental dynamics, as illustrated in Figure~\ref{fig:intro}~(a).
Another line of approaches builds upon video generation models~\cite{epona,GenAD,drivinggpt,drivelaw,Resim,vista}, where intermediate representations are shared between video and action modules through tightly coupled designs, enabling joint modeling of future observations and actions, as illustrated in Figure~\ref{fig:intro}~(b).
Despite achieving impressive performance, existing methods still suffer from two key limitations. VLA-based WAMs require autoregressive future observations generation before action prediction during inference, leading to unavoidable computational latency. Video generation-based WAMs adopt tightly coupled designs, where high-dimensional visual representations can interfere with the low-dimensional action space, resulting in suboptimal trajectory planning.

In this paper, motivated by the observation that embodied navigation ultimately requires producing actions in the physical world, while reasoning and anticipation serve as auxiliary mechanisms to support decision-making, we propose \textbf{\ourMethod}, an end-to-end framework for AD and UN. Our method seamlessly combine action generation and future observations prediction within a WAM framework, as shown in Figure~\ref{fig:intro}~(c). 
Unlike previous methods that jointly model future observations prediction and action generation within a single model, which tightly couples the action prediction and future video generation, such frameworks may inadvertently introduce generative noise into the action space during inference, potentially compromising the precision of action predictions. In contrast, our method decouples the action model from the video generation model, enabling robust action planning without generating future observations.

Specifically, \textbf{\ourMethod} adopt the Mixture-of-Transformers~(MoT) architecture to integrate a video generation expert and an action prediction expert, enabling the modeling of both high-level future scene dynamics and low-level trajectory planning within a shared latent space. This design helps preserve the intrinsic distributional properties of each expert, thereby improving model generalization.
To enable robust and efficient inference, we further introduce an asymmetric attention mask design: future video tokens are allowed to attend to future action tokens, while the reverse direction is masked. This unidirectional information flow enables joint optimization of the video generation and action experts during training, while avoiding explicit future observation generation during inference, and ensuring consistency between training and inference. As a result, our method achieves efficient action planning compared to existing approaches that rely on explicit future prediction.

Our contributions are: 
(\textbf{i})~We propose \textbf{\ourMethod}, an end-to-end World Action Model framework that decouples the action prediction model from the video generation model, enabling each component to maintain its own distributional structure and thereby improving generalization.
(\textbf{ii})~We further introduce an asymmetric attention mask that enforces unidirectional visibility between video and action tokens. This design enables joint training of video and action models, while eliminating the need for explicit video generation during inference and maintaining consistency between training and inference. 
(\textbf{iii})~\textbf{\ourMethod} achieves state-of-the-art performance on the NAVSIM (\texttt{navhard} and \texttt{navtest}) benchmarks for AD and the CityWalker benchmarks for UN. Comprehensive ablation studies confirm the effectiveness of our design, while zero-shot real-world deployments further demonstrate the superior generalization of our approach across diverse environments.

\section{Related work}

\textbf{Video generation models.} 
Recent general video generation models~\cite{sora, wan2025, genie, ali2025world} have advanced rapidly and are increasingly being integrated with downstream tasks such as autonomous driving and navigation. Some works~\cite{doe-1,hu2024drivingworld,vista,drivedreamer,MagicDrive-V2,Gendds,Panacea,MiLA,russell2025gaia} attempt to train autonomous driving video generation action models, primarily focusing on video generation while overlooking the planning objective. 
Some approaches~\cite{drivinggpt,epona,DriveGen,drivelaw} leverage the generative capability of video generation models to improve autonomous driving planning. These methods typically employ diffusion transformer to jointly model and interact with visual and action information.
Some end-to-end approaches~\cite{WoTE,LAW,drive-WM, worlddrive,GenAD,Resim,kirby2026drivor} further incorporate implicit world modeling and reinforcement learning to improve trajectory prediction, modeling scene evolution either in structured representations~\cite{WoTE,seerdrive} or latent space~\cite{LAW,world4drive,worldrft}. 
several navigation methods~\cite{badgr,viking,vint,land} learn policies from expert trajectories, but do not explicitly model future scene evolution. 
However, these methods tightly couple generation and action prediction, which limits generalization.

\textbf{Vision-language-action models.}
Many VLA-based methods~\cite{EMMA,openemma,senna,orion,drivevlm,lmdrive,li2025imagidrive,hermes,automot,omnidrive} leverage the pretrained knowledge of VLMs to achieve strong performance in autonomous driving. Some work~\cite{recogdrive,AutoVLA,alpamayo,openread,AdaThinkDrive} integrates reinforcement learning to enable safe driving via fast–slow reasoning. 
Furthermore, some methods~\cite{pwm,sgdrive,uniugp,fsdrive,unidrivevla,uniworldval,vega} extend vision-language models by incorporating structured prediction of future information. For instance, FSDrive~\cite{fsdrive} and PWM~\cite{pwm} enable future image prediction in large models, while SGDrive~\cite{sgdrive} and DrivePI~\cite{liu2025drivepi} enable occupancy prediction, thereby significantly improving planning quality.
In addition, VLA have also been widely explored in outdoor navigation tasks~\cite{navid,uni-navid,streamvln,internvla-n1,omnivla,RoboTrom-Nav,Navgpt-2,Vlm-social-nav,mcnav}. NavFoM~\cite{navfom} trains a foundation model for navigation and autonomous driving using large-scale data. 
Abot-N0~\cite{abot-n0} unifies navigation tasks in a single training framework and achieves autonomous navigation in urban environments.
In contrast, rather than tightly coupling language and action, our method decouples them and leverages a world model to enhance action modeling, enabling stronger generalization across embodiments and tasks.

\textbf{Worl action models.}
Recent works~\cite{f1-vla,kim2026cosmos,lingbot-va2026,mimic,internvla-a1,openvlaoft,worldvla} in embodied manipulation further leverage a general architecture to integrate multiple experts, including understanding, generation, and action, into a unified model. Building upon this paradigm, some approaches~\cite{fast-wam,dreamzero,GenieEnvisioner,bi2025motus,ye2026world} instantiate WAM, which incorporate world modeling, e.g., predicting future visual states, to support downstream action prediction.
Overall, VLA- and WAM-based paradigms have achieved strong performance in manipulation tasks with relatively static background. However, autonomous driving and navigation involve significantly more complex and dynamic world evolution. In such settings, the WAM paradigm is better suited than VLA to model environmental dynamics and provide richer information for action planning.

\section{Method}

\subsection{Problem formulation and notation}

In autonomous driving~(AD) and urban navigation~(UN), action planning are typically conditioned on the current visual observation~$o_t$ and language instructions~$l$. To enhance action planning robustness, previous WAMs methods~\cite{sgdrive,epona,automot,pwm,lingbot-va2026} incorporate future visual observations and model them jointly with actions during training:
\begin{equation}
p_\theta(a_{t:t+H}, v_{t+1:t+N} \mid o_t, l),
\end{equation}
where $\mathbf{a}_{t:t+H}$ denotes an action chunk with horizon $H$, and $\mathbf{v}_{t+1:t+N}$ represents predicted future observations with horizon $N$. At inference time, these methods either jointly generate actions and future observations via joint denoising,
\begin{equation}
(a_{t:t+H}, v_{t+1:t+N}) \sim 
p_\theta(\cdot \mid o_t, l),
\end{equation}
or adopt an inverse dynamics formulation that infers actions based on explicitly predicted future frames~\cite{pwm,lingbot-va2026}:
\begin{equation}
v_{t+1:t+N} \sim p_\theta(v_{t+1:t+N} \mid o_t, l), \quad
a_{t:t+H} \sim p_\theta(a_{t:t+H} \mid o_t, l, v_{t+1:t+N}).
\end{equation}
However, both paradigms require high-dimensional sampling or recursive denoising of~$v_{t+1:t+N} $ during inference, leading to prohibitive computational overhead and latency.

To address these efficiency bottlenecks, we adopt a decoupled inference paradigm inspired by~\cite{fast-wam}, where future frames are used only as supervision during training, while inference relies solely on the current observation.
Formally, let $z(o_t, l)$ denote the latent representation produced by the video backbone conditioned on the current observation and context. The action prediction is then modeled as:
\begin{equation}
a_{t:t+H} \sim p_\theta(a_{t:t+H} \mid z(o_t, l)).
\end{equation}
In this formulation, future observations are only used during training, while $z(o_t, l)$ is obtained from a single forward pass of the backbone at inference time, enabling efficient real-time planning.

\begin{figure}[t]
    \centering
    \includegraphics[width=\linewidth]{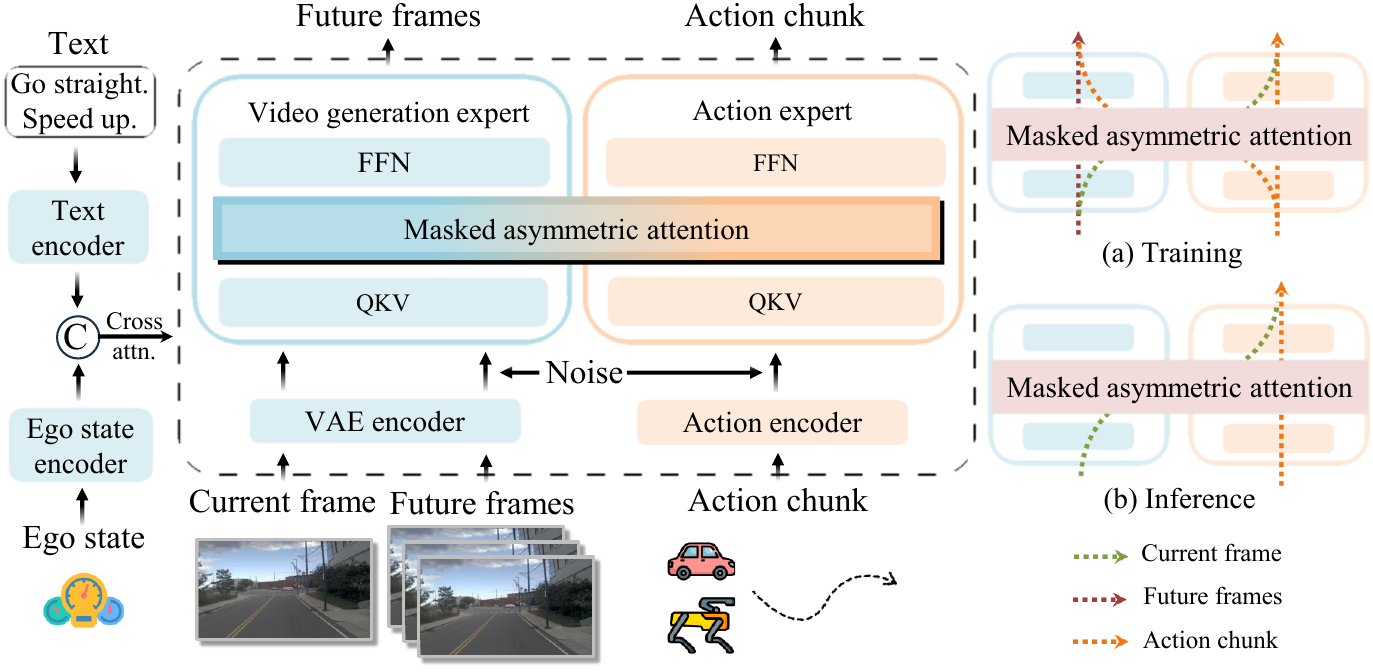}
    \vspace{-5pt}
    \caption{Overview of our \textbf{\ourMethod}. Video and action are jointly learned during training, while inference directly predicts actions from current observation.}
    \label{fig:pipeline}
\end{figure}

\subsection{Network architecture}
To avoid interference between heterogeneous task distributions, we adopt a Mixture-of-Transformers (MoT) architecture to decouple video generation and action planning into two specialized experts,as shown in Figure~\ref{fig:pipeline}. The video generation expert~(VGE) inherits the physical priors from a large-scale video model to capture spatiotemporal dynamics, while the action expert~(AE) is tailored for low-dimensional trajectory prediction. Despite this decoupling, both experts interact through a shared latent space, enabling information exchange while preserving their respective distributional structures.

We adopt Wan2.2-5B~\cite{wan2025} as the backbone of the VGE. We reuse its pre-trained components, including a video VAE for visual encoding and a T5-based text encoder for language conditioning. 
To enable efficient trajectory prediction, our AE is implemented as a diffusion transformer that mirrors the layer depth of the VGE while adopting a smaller model size. In addition, to improve action policy learning, we follow prior work and introduce an agent state encoder to encode the ego state, which is combined with language instructions to condition the model. More details are provided in the Appendix.

Specifically, we organize the model inputs into three types of tokens: latent tokens of the current observation, noisy tokens of future observations, and noisy action tokens for trajectory prediction. All tokens first attend to the language embeddings through cross-attention. They are then projected into a shared latent space via expert-specific projections, where interactions are regulated by a structured attention mechanism. Finally, each expert applies its own feed-forward layers and output heads for task-specific prediction.
This token-level interaction enables controlled information exchange without directly mixing representations across heterogeneous task spaces, thereby preserving the distributional structure within each expert and improving generalization.

\subsection{Structured attention mechanisms.}\label{sec:mask}

Low-latency action planning is a fundamental requirement for both autonomous driving and urban navigation; however, existing WAMs~\cite{pwm,li2025imagidrive,uniworldval,epona} often suffer from heavy inference costs, which severely limits their deployment for fast action prediction. To mitigate this latency, Some WAM approaches~\cite{drivevla,drivelaw} attempt to reduce the prediction horizon, downsample the resolution of future scene forecasting, or avoid explicit future video generation. While these strategies enhance inference speed, they inevitably lead to sub-optimal performance due to the lack of comprehensive spatiotemporal modeling of future states.

\begin{wrapfigure}{r}{0.5\textwidth} 
    \centering
    \includegraphics[width=0.9\linewidth]{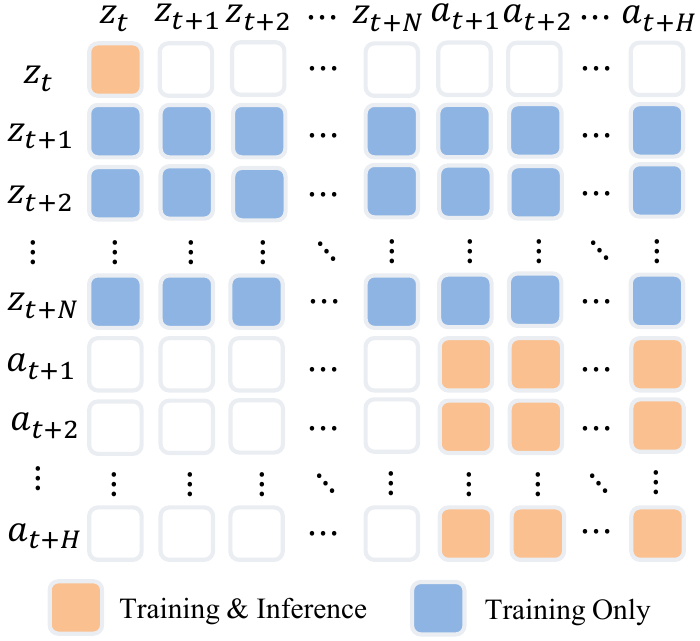} 
    \caption{Asymmetric attention mask: both action and video tokens attend to the current timestep; video tokens can attend to action tokens, but not vice versa.}
    \label{fig:attn_mask}
\end{wrapfigure}

To break this efficiency-accuracy trade-off, we propose a novel asymmetric attention mask within our MoT framework. Unlike previous methods~\cite{epona,drivelaw}, our approach manages the interaction between two experts at the representation level by introducing an asymmetric attention mask that explicitly govern the information flow between action tokens~$a_t$ and latent video tokens~$z_t$, as illustrated in Figure~\ref{fig:attn_mask}. 
Specifically, our design enforces a unilateral visibility constraint: action tokens are restricted to attend only to the current visual observation tokens, ensuring that planning is grounded in the present context. 

In contrast, future video tokens are allowed to attend to both the current observation and all future action tokens.
As shown in Figure~\ref{fig:pipeline}~(b), this structured interaction allows the VGE to predict physically plausible world evolutions conditioned on the specific trajectory intended by the agent. Simultaneously, by sharing the representation space during the joint optimization process, the VGE’s predictive capacity implicitly participates in the action refinement process. This synergy ensures that the AE produces robust trajectories informed by environmental dynamics, while allowing the explicit video generation branch to be completely bypassed during inference to achieve real-time efficiency.

\subsection{Training Objective}\label{sec:train_infer}

We jointly optimize action prediction and future video generation under a flow matching framework. Given current observation~$o_t$ and instruction~$l$, we learn conditional flow models over both action tokens and future video tokens.
For action tokens, we supervise the prediction of the velocity field conditioned on the current context:
\begin{equation}
\mathcal{L}_{\text{act}} =\mathbb{E}_{a_t, \epsilon, t, s} 
\left[
\| u_\theta(a_{t:t+H}^{(s)}, s \mid o_t, l) - \dot{a}_{t:t+H}^{(s)} \|^2
\right],
\end{equation}
where~$s \in [0, 1]$ is the flow time, $a^{(s)}_{t:t+H} = (1 - s)\epsilon + s a_{t:t+H}$, with $\epsilon \sim \mathcal{N}(0, I)$, and the corresponding velocity field is given by~$\dot{a}_{t:t+H} = a_{t:t+H} - \epsilon$.
For future video tokens, we supervise the prediction of velocity of field conditioned on the current context and action tokens:
\begin{equation}
\mathcal{L}_{\text{video}} =
\mathbb{E}\left[\| u_\phi(z^{(s)}_{t+1:t+H}, s \mid o_t, \hat{a}_{t:t+H},l) - \dot{z}^{(s)}_{t+1:t+H} \|^2 \right], 
\end{equation}
where $\hat{a}_{t:t+H}$ denotes the predicted future action sequence, $z^{(s)}_{t+1:t+H} = (1 - s)\epsilon + s z_{t+1:t+H}$. The overall training objective is:~$\mathcal{L} = \mathcal{L}_{\text{action}} + \lambda \mathcal{L}_{\text{video}}$ where~$\lambda=1$ is a weighting coefficient that balances action learning and world modeling.

\section{Experiments}

\textbf{Implementation details.}  
We adopt Wan2.2-5B as the VGE and the AE shares the same architecture with a reduced hidden dimension~($d_a = 1024$), resulting in a 1B-parameter branch and an overall model size of 6B.
For task setup, we use an action horizon of 8 for NAVSIMv2~(4 seconds with 0.5-second intervals), where each waypoint is represented as~$(x, y, \theta)$, and a horizon of 5 for CityWalker with~$(x, y)$. We use only the front-facing camera as input, and align video and action chunks with a 1:1 temporal ratio.
Both VGE and AE are trained under the same flow matching formulation. For NAVSIMv2, we use an input resolution of~$640 \times 768$ and train for 60 epochs with a batch size of 64; for CityWalker, we use~$384 \times 384$ and train for 30 epochs with the same batch size.
During inference, we use 10 denoising steps with classifier-free guidance~(CFG = 1.0). All experiments are conducted on 8 NVIDIA H200 GPUs.

\begin{table*}[t]
\centering
\caption{\textbf{Performance on the NAVSIM-v2 \texttt{navhard} Leaderboard.} PDM-Closed uses ground-truth symbolic inputs for planning. \dag~ indicates values are copied from~\cite{simscale} and * indicates values are copied from~\cite{guideflow}; all other results are reproduced with the official code repository or official checkpoints. (S.: per-stage EPDM score.)}
\label{tab:navhard}
\vspace{-5pt}
\renewcommand\arraystretch{1.0}
\setlength{\tabcolsep}{1.0mm}
\resizebox{\linewidth}{!}{
\begin{tabular}{l|l|c|cccc|ccccc|c|c}
\toprule
Method & Reference & Stage 
& NC $\uparrow$ & DAC $\uparrow$ & DDC $\uparrow$ & TLC $\uparrow$ 
& EP $\uparrow$ & TTC $\uparrow$ & LK $\uparrow$ 
& HC $\uparrow$ & EC $\uparrow$ 
& \textbf{S.} $\uparrow$
& EPDMS $\uparrow$ \\
\midrule

\multirow{2}{*}{PDM-Closed~\cite{dauner2023parting}} 
& \multirow{2}{*}{-} 
& S1 & 94.4 & 78.8 & 100 & 99.5 & 100 & 93.5 & 99.3 & 87.7 & 36.0 & - &  \\
& & S2 & 88.1 & 90.6 & 96.3 & 98.5 & 100 & 83.1 & 73.7 & 91.5 & 25.4 & - & \multirow{-2}{*}{51.3} \\ 
\cmidrule{1-14}

\rowcolor[HTML]{EEE5F4}\hline
\multicolumn{14}{c}{\textit{{Traditional E2E-based Planner}}}\\

\multirow{2}{*}{LTF\dag~\cite{transfuser}} 
& \multirow{2}{*}{T-PAMI2022} 
& S1 & 97.3 & 80.2 & 97.8 & 99.3 & 83.4 & 96.2 & 92.9 & 97.8 & 71.1 & 61.3 & \cellcolor{gray!20} \\
& & S2 & 79.4 & 69.0 & 85.6 & 98.5 & 83.8 & 76.7 & 47.9 & 97.0 & 70.6 & 39.2 & \multirow{-2}{*}{\cellcolor{gray!20}24.4} \\ 
\cmidrule{1-14}

\multirow{2}{*}{DiffusionDrive\dag~\cite{diffusiondrive}} 
& \multirow{2}{*}{CVPR2025} 
& S1 & 96.8 & 86.0 & 98.8 & 99.3 & 84.0 & 95.8 & 96.7 & 97.6 & 79.6 & 66.7 & \cellcolor{gray!20} \\
& & S2 & 80.1 & 72.8 & 84.4 & 98.4 & 85.9 & 76.6 & 46.4 & 96.3 & 72.8 & 40.5 & \multirow{-2}{*}{\cellcolor{gray!20}27.5} \\ 
\cmidrule{1-14}

\multirow{2}{*}{GTRS-DP*~\cite{gtrs}} 
& \multirow{2}{*}{CVPRW2025} 
& S1 & 94.7 & 78.8 & 96.1 & 99.5 & 83.0 & 94.4 & 92.0 & 97.5 & 72.8 & - & \cellcolor{gray!20} \\
& & S2 & 80.3 & 74.4 & 84.9 & 98.0 & 81.9 & 78.8 & 45.4 & 96.7 & 70.1 & - & \multirow{-2}{*}{\cellcolor{gray!20}23.8} \\ 
\cmidrule{1-14}

\multirow{2}{*}{GuideFlow*~\cite{guideflow}} 
& \multirow{2}{*}{CVPR2026} 
& S1 & 96.6 & 80.5 & 96.3 & 99.3 & 82.3 & 94.9 & 91.5 & 97.7 & 67.8 & - & \cellcolor{gray!20} \\
& & S2 & 87.3 & 76.7 & 88.8 & 99.2 & 84.3 & 85.1 & 49.7 & 93.1 & 44.5 & - & \multirow{-2}{*}{\cellcolor{gray!20}27.1} \\ 

\rowcolor[HTML]{EEE5F4}\hline
\multicolumn{14}{c}{\textit{{VLA-based Planner}}}\\

\multirow{2}{*}{ReCogDrive~\cite{recogdrive}} 
& \multirow{2}{*}{ICLR2026} 
& S1 & 96.4 & 78.9 & 98.7 & 99.8 & 82.6 & 95.6 & 94.4 & 97.6 & 74.2 & 67.7 & \cellcolor{gray!20} \\
& & S2 & 80.2 & 65.0 & 82.4 & 98.7 & 85.2 & 76.9 & 43.8 & 96.6 & 71.8 & 37.6 & \multirow{-2}{*}{\cellcolor{gray!20}25.7} \\ 
\cmidrule{1-14}

\multirow{2}{*}{SGDrive~\cite{sgdrive}} 
& \multirow{2}{*}{CVPR2026} 
& S1 & 95.8 & 87.6 & 97.8 & 99.8 & 84.4 & 94.7 & 92.9 & 97.8 & 28.9 & 71.1 & \cellcolor{gray!20} \\
& & S2 & 79.4 & 65.4 & 79.1 & 98.9 & 88.9 & 75.3 & 42.7 & 96.4 & 29.6 & 35.2 & \multirow{-2}{*}{\cellcolor{gray!20}25.5} \\ 



\rowcolor[HTML]{EEE5F4}\hline
\multicolumn{14}{c}{\textit{{WAM-based Planner}}}\\


\multirow{2}{*}{\ourMethod~(Ours)} 
& \multirow{2}{*}{-} 
& S1 & 96.6 & 87.8 & 99.0 & 99.3 & 84.5 & 95.6 & 97.8 & 97.8 & 77.8 & \textbf{75.8} & \cellcolor{gray!20} \\
& & S2 & 79.6 & 73.3 & 84.9 & 97.8 & 85.8 & 76.6 & 47.7 & 95.4 & 75.3 & \textbf{41.7} & \multirow{-2}{*}{\textbf{\cellcolor{gray!20}32.2}} \\ 

\bottomrule
\end{tabular}
}
\vspace{-5pt}
\end{table*}

We evaluate our method on two real-world datasets covering autonomous driving and urban navigation scenarios, namely NAVSIM-v2 and CityWalk. Detailed dataset descriptions and evaluation metrics are provided in the Appendix.

\textbf{NAVSIM-v2.}
We train our model on the navtrain subset (1,192 scenarios) and evaluate performance on two benchmarks: \texttt{navhard} and \texttt{navtest}~\cite{NAVSIMv2}. \texttt{navhard} comprises 244 safety-critical real-world scenarios (Stage 1) and 4,164 3DGS-generated synthetic counterparts (Stage 2) for closed-loop evaluation. \texttt{navtest} includes 12,146 scenarios for assessing generalization.
The two benchmarks share a rule-based planning metric, EPDMS, which evaluates performance using multiple sub-metrics, including No-at-fault Collisions~(NC), Drivable Area Compliance~(DAC), Driving Direction Compliance~(DDC), Traffic Light Compliance~(TLC), Time-to-Collision~(TTC), Ego Progress~(EP), Lane Keeping~(LK), History Comfort~(HC), and Extended Comfort~(EC).
We also provide results on NAVSIM-v1 \texttt{navtest} with PDMS in the Appendix.

\textbf{CityWalker.}
The dataset~\cite{citywalker} contains 15 hours of teleoperation data collected across diverse urban areas in New York City, with 6 hours used for fine-tuning and 9 hours for testing.
We evaluate performance under several challenging scenarios, including turning, intersection crossing, detours, proximity to pedestrians, and crowded environments. 
We utilize Maximum Average Orientation Error~(MAOE) as the primary metric to assess human-like trajectory alignment, supplemented by the average L2 distance for spatial accuracy.

%
\textbf{Real-world experiments}
We deploy our model and the baseline methods on the same Unitree Go2 quadruped for real-world navigation. We adapt the PD controller provided in ~\cite{internvla-n1} to provide velocity command to the quadruped. We conduct zero-shot real-world experiments in both indoor and outdoor environments, demonstrating the strong generalization ability of our \textbf{\ourMethod}. Detailed qualitative results are provided in the appendix, and additional videos are included in the supplementary material.

\subsection{Main results}

\begin{table*}[th]
\centering
\caption{\textbf{Performance comparison on NAVSIM-v2} \texttt{navtest} \textbf{Leaderboard}. 
* indicates training with reinforcement learning; 
$\dag$ indicates training on the full navtrain split; 
$\ddagger$ indicates the use of the best-of-$N$ ($N=6$) strategy following~\cite{AutoVLA}.}
\label{tab:main_results_on_epdms_navtest}
\vspace{-5pt}
\resizebox{\linewidth}{!}{
\begin{tabular}{l|c|cccc|ccccc|c}
\toprule
Method & Sensors & NC$\uparrow$ & DAC$\uparrow$ & DDC$\uparrow$ & TLC$\uparrow$ & EP$\uparrow$ & TTC$\uparrow$ & LK$\uparrow$ & HC$\uparrow$ & EC$\uparrow$ & EPDMS$\uparrow$ \\
\midrule
Human Agent & - & 100 & 100 & 99.8 & 100 & 87.4 & 100 & 100 & 98.1 & 90.1 & 90.3 \\
\midrule

\rowcolor[HTML]{EEE5F4}\hline
\multicolumn{12}{c}{\textit{Traditional E2E-based Planner}}\\

TransFuser~\cite{transfuser} & 3xC+L & 96.9 & 89.9 & 97.8 & 99.7 & 87.1 & 95.4 & 92.7 & 98.3 & 87.2 & \cellcolor{gray!20}76.7 \\
Hydra-MDP++~\cite{Hydra-mdp++} & 3xC+L  & 97.2 & 97.5 & 99.4 & 99.6 & 83.1 & 96.5 & 94.4 & 98.2 & 70.9 & \cellcolor{gray!20}81.4 \\
GTRS-Dense~\cite{gtrs} & 3xC  & 97.6 & 97.5 & 99.0 & 99.9 & 87.9 & 97.0 & 95.9 & 97.5 & 55.9 & \cellcolor{gray!20}82.3 \\
DriveSuprim~\cite{drivesuprim} & 3xC  & 97.5 & 96.5 & 99.4 & 99.6 & 88.4 & 96.6 & 95.5 & 98.3 & 77.0 & \cellcolor{gray!20}83.1 \\
ARTEMIS~\cite{artemis} & 3xC+L  & 98.3 & 95.1 & 98.6 & 99.8 & 81.5 & 97.4 & 96.5 & 98.3 & - & \cellcolor{gray!20}83.1 \\
DiffusionDrive~\cite{diffusiondrive} & 3xC+L  & 98.2 & 95.9 & 99.4 & 99.8 & 87.5 & 97.3 & 96.8 & 98.3 & 87.7 & \cellcolor{gray!20}84.5 \\
World4Drive~\cite{world4drive} & 3xC & 97.8 & 96.3 & 99.4 & 99.8 & 88.3 & 97.1 & 97.7 & 98.0 & 53.9 & \cellcolor{gray!20}84.8 \\
Drive-JEPA~\cite{drive-jepa} & 1xC & 98.8 & 97.4 & 99.0 & 99.8 & 83.5 & 98.0 & 96.2 & 98.1 & 85.6 & \cellcolor{gray!20}85.4 \\

WorldRFT*~\cite{worldrft} & 3xC & 97.8 & 96.5 & 99.5 & 99.8 & \textbf{88.5} & 97.0 & 97.4 & 98.1 & 69.1 & \cellcolor{gray!20}86.7 \\

\rowcolor[HTML]{EEE5F4}\hline
\multicolumn{12}{c}{\textit{VLA-based Planner}}\\

ReCogDrive*~\cite{recogdrive} & 1xC & 98.3 & 95.2 & 99.5 & 99.8 & 87.1 & 97.5 & 96.6 & 98.3 & 86.5 & \cellcolor{gray!20}83.6 \\
SGDrive~\cite{sgdrive} & 1xC & 98.6 & 94.3 & 99.5 & \textbf{99.9} & 86.0 & 97.9 & 96.1 & 98.3 & 85.9 & \cellcolor{gray!20}86.2 \\

Vega~\cite{vega} & 1xC & \textbf{98.9} & 95.3 & 99.4 & \textbf{99.9} & 87.0 & \textbf{98.4} & 96.1 & 98.3 & 76.3 & \cellcolor{gray!20}86.9 \\
DriveFine*~\cite{drivefine} & 1xC & 98.7 & 97.3 & 98.8 & 99.8 & 88.2 & 97.8 & 97.7 & \textbf{98.4} & 84.7 & \cellcolor{gray!20}87.1 \\

\rowcolor[HTML]{EEE5F4}\hline
\multicolumn{12}{c}{\textit{WAM-based Planner}}\\

Epona~\cite{epona} & 1xC & 97.1 & 95.7 & 99.3 & 99.7 & \textbf{88.6} & 96.3 & 97.0 & 98.0 & 67.8 & \cellcolor{gray!20}85.1 \\
DriveVLA-W0~\dag~\cite{drivevla} & 1xC & 98.5 & \textbf{99.1} & 98.0 & 99.7 & 86.4 & 98.1 & 93.2 & 97.9 & 58.9 & \cellcolor{gray!20}86.1 \\

\rowcolor{gray!20}\textbf{\ourMethod~(Ours)} & 1xC & 98.4 & 97.2 & \textbf{99.6} & 99.8 & 87.8 & 97.7 & 97.8 & \textbf{98.4} & 88.0 & 89.5 \\
\rowcolor{gray!20}\textbf{\ourMethod$~\ddagger$~(Ours)} & 1xC & 98.5 & 97.5 & \textbf{99.6} & 99.8 & 87.9 & 97.8 & \textbf{98.0} & \textbf{98.4} & \textbf{90.0} & \textbf{90.3} \\

\bottomrule
\end{tabular}}
\vspace{-5pt}
\end{table*}

\textbf{Navhard Leaderboard.} 
We first evaluate our method in safety-critical closed-loop scenarios. As shown in Table~\ref{tab:navhard}, our method achieves the best performance in both Stage 1 and Stage 2, consistently outperforming prior approaches. Overall, it attains an EPDMS score of 32.2, demonstrating the effectiveness of the WAM paradigm in producing smooth and comfortable driving behaviors.
Notably, compared to VLA-based methods~\cite{recogdrive,sgdrive}, our approach achieves consistent improvements on DAC and DDC across both stages, which measure rule compliance and trajectory feasibility. In particular, it surpasses prior methods on Stage 2 by at least 7.9 on DAC and 2.5 on DDC. This indicates that, in the absence of explicit map supervision, world modeling provides a stronger inductive bias for learning environment-constrained driving behaviors. 
In contrast, VLA-based methods, which rely on scene understanding and high-level reasoning, are less effective at enforcing such constraints during action generation.

\begin{table*}[t]
    \centering
    \caption{\textbf{Performance comparison on CityWalker dataset}. Percentages indicate the proportion of data in each scenario. “Mean” denotes the average performance across the six scenarios, while “All” represents the average over all samples. The best results are highlighted in bold. * indicates a pretrained method using large-scale navigation data.}
    \label{tab:citywalk}
    \vspace{-5pt}
    \resizebox{0.9\linewidth}{!}{
    \begin{tabular}{l|l|cccccccc}
    \toprule
        \multirow{2}{*}{\textbf{Method}} & \multirow{2}{*}{\textbf{Metric}} & \multirow{2}{*}{\textbf{Mean}} & \textbf{Turn} & \textbf{Crossing} & \textbf{Detour} & \textbf{Proximity} & \textbf{Crowd} & \textbf{Other} & \textbf{All} \\ 
        & & & 8\% & 12\% & 12\% & 6\% & 7\% & 55\% & 100\% \\ 
        \midrule

        \rowcolor[HTML]{EEE5F4}\hline
        \multicolumn{10}{c}{\textit{{Pretrained method}}}\\
        
        \multirow{2}{6em}{ABot-N0*~\cite{abot-n0}} 
         & $\downarrow$ L2 (m) & - & - & - & - & - & - & - & -\\ 
         & $\downarrow$ MAOE ($^{\circ}$) & \textbf{11.2} & 21.3 & \textbf{9.8} & 12.8 & \textbf{8.1} & 8.8 & \textbf{6.3} & \textbf{7.6}\\ 
        
        \rowcolor[HTML]{EEE5F4}\hline
        \multicolumn{10}{c}{\textit{{Fine-tuned method}}}\\
        
        \multirow{2}{6em}{GNM~\cite{gnm}} 
         & $\downarrow$ L2 (m) & 1.22 & 2.36 & 1.36 & 1.42 & 0.88 & 0.76 & \textbf{0.55} & 0.74\\
         & $\downarrow$ MAOE ($^{\circ}$) & 16.2 & 31.1 & 14.8 & 12.5 & 14.7 & 12.8 & 11.0 & 12.1\\
        \midrule
        \multirow{2}{6em}{ViNT~\cite{vint}} 
         & $\downarrow$ L2 (m) & 1.30 & 1.91 & 1.13 & 1.14 & 0.77 & \textbf{0.66} & 0.57 & 0.70\\
         & $\downarrow$ MAOE ($^{\circ}$) & 16.5 & 31.1 & 15.4 & 12.9 & 14.8 & 13.3 & 11.6 & 12.6\\
        \midrule
        \multirow{2}{6em}{NoMaD~\cite{nomad}} 
         & $\downarrow$ L2 (m) & 1.39 & 2.49 & 1.56 & 1.55 & 1.06 & 0.95 & 0.76 & 0.74\\
         & $\downarrow$ MAOE ($^{\circ}$) & 19.1 & 35.1 & 18.5 & 15.6 & 18.1 & 14.3 & 12.8 & 12.1\\
        \midrule
        \multirow{2}{6em}{CityWalker~\cite{citywalker}} 
         & $\downarrow$ L2 (m) & 1.11 & 1.27 & 1.00 & 1.15 & 1.06 & 1.12 & 1.06 & 1.07\\ 
         & $\downarrow$ MAOE ($^{\circ}$) & 15.2 & 26.6 & 14.1 & 13.9 & 14.3 & 12.0 & 10.4 & 11.5\\ 
        \midrule
         
        \multirow{2}{6em}{\ourMethod~(Ours)} 
         & $\downarrow$ L2 (m) & \textbf{0.71} & \textbf{0.69} & \textbf{0.77} & \textbf{0.66} & \textbf{0.76} & 0.77 & 0.63 & \textbf{0.64}\\ 
         & $\downarrow$ MAOE ($^{\circ}$) & 11.8 & \textbf{19.5} & 11.9 &\textbf{9.1} & 11.4 & \textbf{8.0} & 7.8 & 9.8\\ 
    \bottomrule
    \end{tabular}
    \vspace{-5pt}
}
\end{table*}
\textbf{Navtest Leaderboard.}
On the more diverse \texttt{navtest} benchmark,as shown in Table~\ref{tab:main_results_on_epdms_navtest}, \textbf{\ourMethod}~achieves the state-of-the-art EPDMS score under a fair comparison setting, notably without relying on multi-stage training, reinforcement learning, or auxiliary datasets. It surpasses prior VLA-based methods by at least 2.4 points, maintaining competitive or superior performance across all metrics. This consistent advantage underscores the efficacy of integrating world modeling into action planning.
Specifically, compared to methods that predict future states at fixed timestamps~\cite{vega, sgdrive}, \ourMethod~shows significant gains in EP, LK, and EC. This demonstrates that by internalizing physically grounded dynamics during training, the action expert generates more rule-compliant and comfortable behaviors. Notably, even without RL-based fine-tuning, our approach maintains a clear advantage in EC over models~\cite{recogdrive, drivefine} specifically optimized for such metrics, further proving the robustness of our WAM-based representation.

\textbf{CityWalker dataset.}
Compared to fine-tuning-based methods,as shown in Table~\ref{tab:citywalk}, \ourMethod achieves leading performance in both L2 error and MAOE, validating the superiority of the WAM paradigm. When compared with ABot-N0~\cite{abot-n0} pretrained on large-scale datasets, our approach exhibits lower MAOE in complex scenarios such as Turn Detour and Crowd navigation. This demonstrates that integrating latent world dynamics with action planning provides more effective guidance than relying on logical reasoning within the linguistic space. In dense urban environments, the ability to model spatiotemporal evolution allows the AE to predict more plausible, collision-free trajectories, overcoming the limitations of language representations in providing fine-grained motion guidance.
\begin{figure}[t]
    \centering
    \includegraphics[width=1\linewidth]{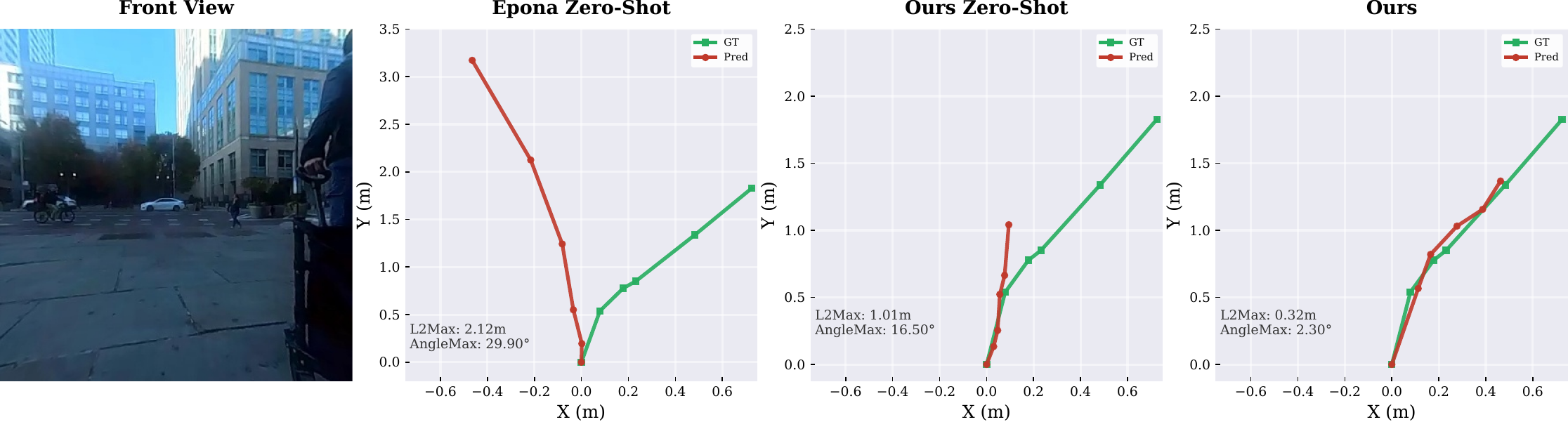}
    \vspace{-15pt}
    \caption{Qualitative results on CityWalker. We present the zero-shot results of Epona and our method (\ourMethod), as well as the fine-tuned results of our method.}
    \label{fig:cw_vis}
    \vspace{-15pt}
\end{figure}

\begin{table}[t] 
    \centering
    \begin{minipage}[b]{0.53\linewidth}
        \centering
        \caption{Ablation of image size and attention mask.}
        \label{tab:combined_ablation}
        \resizebox{\linewidth}{!}{
            \begin{tabular}{llcccc}
            \toprule
            \multirow{2}{*}{\textbf{Image Size}} & \multirow{2}{*}{\textbf{variant}} & \multicolumn{3}{c}{\textbf{navtest} $\uparrow$} & \textbf{navhard} $\uparrow$ \\
            \cmidrule(lr){3-5} \cmidrule(lr){6-6}
            & & DAC & LK & EPDMS & EPDMS \\
            \midrule
            \multirow{3}{*}{320$\times$384} & Joint     & 96.5 & 97.0 & 87.4 & 28.0 \\
                                             & Isolated & 96.5 & 97.5 & 88.3 & 29.4 \\
                                             & Ours      & 97.0 & 97.6 & 88.8 & 31.6 \\
            \midrule
            640$\times$768                   & Ours      & \textbf{97.5} & \textbf{98.0} & \textbf{89.5} & \textbf{32.2} \\
            \bottomrule
            \end{tabular}
        }
    \end{minipage}
    \hfill 
    \begin{minipage}[b]{0.46\linewidth}
        \centering
        \caption{Ablation of denoising steps.}
        \label{tab:step}
        \resizebox{0.84\linewidth}{!}{
            \begin{tabular}{c|c|ccc}
            \toprule
            \multirow{2}{*}{\textbf{Steps}} & \textbf{navtest} & \multicolumn{3}{c}{\textbf{navhard}} \\
            \cmidrule(lr){3-5}
            & EPDMS & S1 & S2 & EPDMS \\
            \midrule
            1  & 87.2 & 74.5 & 39.8 & 30.4 \\
            2  & 89.2 & 76.0 & 40.7 & 31.2 \\
            5  & 89.4 & 74.5 & 41.4 & 31.4 \\
            10 & \textbf{89.5} & \textbf{75.8} & \textbf{41.7} & \textbf{32.2} \\
            \bottomrule
            \end{tabular}
        }
    \end{minipage}
\end{table}
\subsection{Ablation studies}

\textbf{Ablation of asymmetric attention mask.}
To evaluate the effect of the proposed asymmetric attention mask, we construct two variants. The first is joint attention, which is commonly adopted in VLA-based and video-generation-based WAMs, where future video and action tokens are fully coupled. The second is isolated attention, where the two experts are completely decoupled and future video and action tokens are mutually invisible.
As shown in Table~\ref{tab:combined_ablation}, under the same input resolution, all variants achieve competitive performance. Our asymmetric attention mask consistently yields the best results, with more pronounced improvements on challenging scenarios.
Compared to joint attention, the significant performance gain highlights the importance of decoupling future video generation from action prediction, as it helps preserve the stability of the action space. In contrast, compared to isolated attention, our method enables the action expert to implicitly capture future dynamics during training, leading to improved planning performance, particularly in complex scenarios. 
More detailed analyses are provided in the appendix~\ref{discussion}.

\textbf{Image size.}
We study the impact of input image resolution on our method, as shown in Table~\ref{tab:combined_ablation}. Reducing the resolution leads to consistent performance degradation on both \texttt{navtest} and \texttt{navhard}. In particular, the EPDMS score drops by 0.7 and 0.6 points on \texttt{navtest} and \texttt{navhard}, respectively. 
We also observe noticeable declines in DAC and LK, showing that higher resolution inputs provide richer spatial and geometric cues, which facilitate the action expert in capturing scene dynamics and producing more feasible and stable trajectories.

\textbf{Denoising steps.}
We analyze the impact of different numbers of denoising steps on performance. As shown in Table~\ref{tab:denoise_step}, increasing the number of steps consistently improves performance on both \texttt{navtest} and \texttt{navhard}. 
While the performance gain is relatively modest in general scenarios, it becomes more pronounced on challenging benchmarks, indicating that additional denoising steps help refine predictions under complex dynamics. 


\textbf{Ablation of different VGE and AE.}
The impact of different VGE backbones is investigated at a $640 \times 768$ resolution, as shown in Table~\ref{tab:vge_ae}. Results indicate that although the Wan2.1-1.3B~\cite{wan2025} variant achieves comparable performance on \texttt{navtest}, its deficiency in \texttt{navhard} reveals the limitations of lower-capacity generative priors in complex scenarios. We hypothesize that a more powerful VGE serves as a more robust world model, implicitly bolstering the AE’s reasoning via shared latent representations. Simultaneously, increasing the AE size under a fixed VGE further improves generalization across all benchmarks, underscoring the benefits of scaling the policy head alongside the generative backbone.

\textbf{Inference latency of different methods.}
We compare the inference latency of our \textbf{\ourMethod}~with several representative approaches, as shown in Table~\ref{tab:latency}. 
Epona and PWM represent WAM methods based on video generation models and large vision-language models, respectively. All methods are evaluated on a single NVIDIA RTX 4090 GPU.
The configurations are determined by practical requirements. The video generation setting uses 10 denoising steps to ensure high-quality future prediction, while the action only setting uses only 2 denoising steps for efficient action inference while maintaining accuracy.
We also report PDMS and EPDMS on \texttt{navtest} (full results are provided in the Appendix). Compared to the variant that generates future video, our action-only inference achieves up to $8\times$ speedup, demonstrating the efficiency of our approach.

\begin{figure}[t]
    \centering
    \includegraphics[width=1\linewidth]{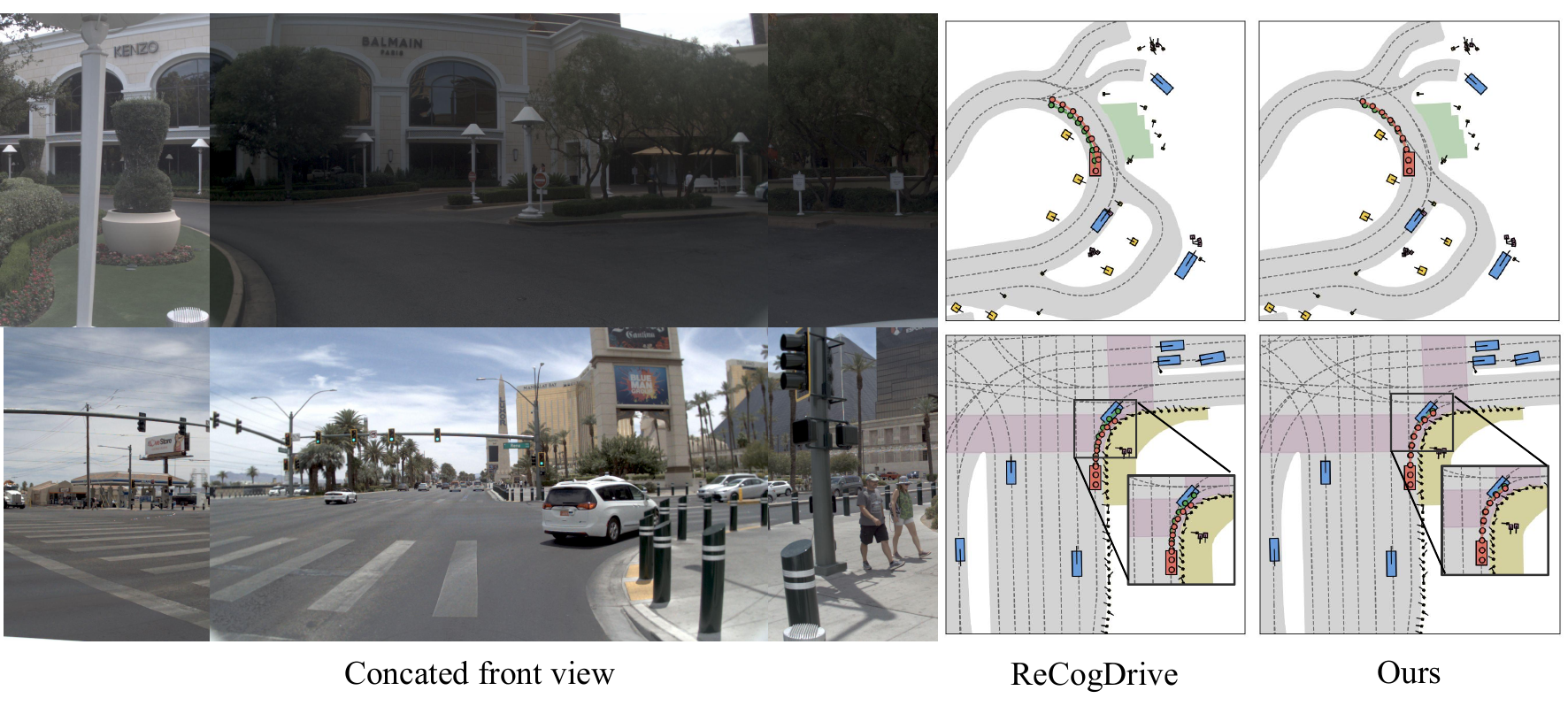}
    \vspace{-15pt}
    \caption{Qualitative evaluation of turning scenarios on NAVSIM. 
    }
    \label{fig:nav_vis}
    \vspace{-10pt}
\end{figure}

\begin{figure}[t]
    \centering
    \begin{minipage}[t]{0.48\linewidth}
        \centering
        \captionof{table}{Ablation of different VGE and AE.}
        \vspace{-5pt}
        \label{tab:vge_ae}
        \resizebox{\linewidth}{!}{
        \begin{tabular}{l|c|cc|c}
        \toprule
        \multirow{2.5}{*}{VGE} & \multirow{2.5}{*}{AE} & \multicolumn{2}{c|}{\textbf{navtest} $\uparrow$} & \textbf{navhard} $\uparrow$ \\
        \cmidrule(lr){3-4} \cmidrule(lr){5-5}
        & & PDMS & EPDMS & EPDMS \\
        \midrule
        Wan2.1-1.3B & $\sim$0.24B  & 88.5 & 88.8 & 28.8 \\
        Wan2.2-14B  & $\sim$0.21B  & 88.2 & 88.2 & 31.2 \\
        Wan2.2-14B  & $\sim$1.04B  & \textbf{89.1} & \textbf{89.5} & \textbf{32.2} \\
        \bottomrule
        \end{tabular}
        }

    \end{minipage}
    \hfill 
    \begin{minipage}[t]{0.48\linewidth}
        \centering
        \captionof{table}{Inference latency with other methods.}
        \label{tab:latency}
        \vspace{-5pt}
        \resizebox{0.95\linewidth}{!}{
            \begin{tabular}{c|ccc}
            \toprule
            Method & PDMS  &EPDMS & latency(s)\\
            \midrule
            Epona~\cite{epona} & 86.2 & 85.1 &  0.32 \\
            PWM~\cite{pwm} & 87.3 & - & 0.57 \\
            PWM~(w/ video)~\cite{pwm} & 88.1 & - & 0.83 \\
            \ourMethod~(w/ video)~(ours) & \textbf{89.0} & \textbf{89.5} & 1.38 \\
            \ourMethod~(ours) & 88.9 & 89.2& \textbf{0.17}\\
            \bottomrule
            \end{tabular}
        }
    \end{minipage}
    \vspace{-5pt}
\end{figure}

\subsection{Qualitative Analysis}

We provide qualitative results on both UN and AD tasks. 
For the UN task, as shown in Figure~\ref{fig:cw_vis}, our method demonstrates stronger generalization compared to Epona, achieving better zero-shot performance. After fine-tuning, it produces more accurate and stable navigation trajectories.
For the AD task, compared with the previous state-of-the-art method ReCogDrive~\cite{recogdrive}, our approach exhibits more reasonable and smoother behaviors, particularly in turning scenarios, as shown in Figure~\ref{fig:nav_vis}. The predicted trajectories better align with the underlying scene geometry and driving constraints.
These results validate that our method effectively improves trajectory planning by enabling the action expert to implicitly capture scene dynamics during training.

\section{Conclusion}
In this paper, we propose \textbf{\ourMethod}, an end-to-end WAM framework for AD and UN built upon a Mixture-of-Transformers architecture. Distinct from prior WAMs, \textbf{\ourMethod} introduces an asymmetric attention mask mechanism that enables a unique "joint training, decoupled inference" paradigm. While we jointly optimize future video generation and action prediction during training to capture world dynamics, our design allows for efficient action planning without the need for explicit future observation synthesis during inference. By decoupling the low-dimensional action space from high-dimensional visual generation, we preserve the distributional integrity and stability of the action model, significantly enhancing both generalization and inference efficiency. Extensive evaluations on autonomous driving and urban navigation benchmarks demonstrate that our approach achieves state-of-the-art performance, validating the effectiveness of our decoupled WAM design.

\clearpage
\newpage
{
    \small
    \bibliographystyle{plain}
    \bibliography{main}
}


\newpage
\appendix
\section*{Appendix}
\startcontents[appendices]
\printcontents[appendices]{}{1}{\section*{\contentsname}\vskip -2ex}
\vspace{1cm}
\hrule

\section{Discussion}\label{discussion}

\subsection{Motivation}
In this study, we explore the relationship between video generation and action prediction within World action models~(WAMs) for autonomous driving and urban navigation. Drawing inspiration from human cognitive processes as discussed in our introduction, we propose a framework characterized by joint training and decoupled inference. During the training phase, a loose coupling mechanism—implemented via a simple asymmetric attention module—ensures that the video generation expert~(VGE) generate a future consistent with the ego-actions. This connection allows the gradients from the generation expert to backpropagate into the action expert~(AE), implicitly optimizing the latter and ensuring a stable distribution within the action space. In contrast, during inference, the generation process is entirely bypassed; the model performs efficient decision-making by leveraging only the current observations for contextual information, thereby eliminating the need for explicit future world generation.

\subsection{Discussion about different attention masks}
Based on the experimental results presented in the Table~\ref{tab:dis_abl}, our asymmetric attention mask achieves superior performance across all metrics under identical input conditions.

While common joint training methods~(represented by Joint) exhibit strong overall performance relative to baseline approaches~\cite{pwm,drivelaw,uniworldval,epona}, they lag significantly in DAC and EP metrics. This performance gap stems from the fact that during inference, the action space in these models is heavily injected with noise from the generation process, which interferes with the integrity of action prediction. As illustrated in Table 2, our method maintains an absolute lead in EP and DAC among WAM-based approaches, achieving a 2.7\% improvement even over DriveLaW~\cite{drivelaw}, the highest-rated model overall.

Regarding the Isolated attention mask, the complete separation of tasks prevents the action space from effectively utilizing the visual context provided by current observations. Although its DAC and EP scores are comparable to ours, its lower overall rating underscores the necessity of implicitly optimizing the action expert through the generation expert during training. This balance ensures that the model leverages sophisticated world-understanding without suffering from inference-time generation noise.

\begin{table}[ht]
    \centering
    \caption{Detailed discussion on attention mask variants.}
    \label{tab:dis_abl}
    \resizebox{\linewidth}{!}{
        \begin{tabular}{llccccccccc}
            \toprule
            \multirow{2}{*}{\textbf{Image Size}} & \multirow{2}{*}{\textbf{Variant}} & \multicolumn{3}{c}{\textbf{navtest} $\uparrow$} & \multicolumn{3}{c}{\textbf{navtest} $\uparrow$} & \textbf{navhard} $\uparrow$ \\
            \cmidrule(lr){3-5} \cmidrule(lr){6-8} \cmidrule(lr){9-9}
            & & DAC & EP & PDMS & DAC & EP & EPDMS & EPDMS\\ 
            \midrule
            \multirow{3}{*}{320$\times$384} & Joint    &95.7 & 81.6& 87.1& 96.5 & 87.6 & 87.4 &  28.0\\
                                            & Isolated & 96.5 & 82.6 & 88.0 &96.9 & 87.7 & 88.3 &  29.4\\
                                            & Ours  & \textbf{96.9} & \textbf{82.9} & \textbf{88.3}  & \textbf{97.0} & \textbf{87.8} & \textbf{88.8} & \textbf{31.6} \\
            \bottomrule
        \end{tabular}
    }
\end{table}
\section{Real world experiments}

\begin{figure}[t]
    \vspace{-15pt}
    \centering
    \includegraphics[width=0.9\linewidth]{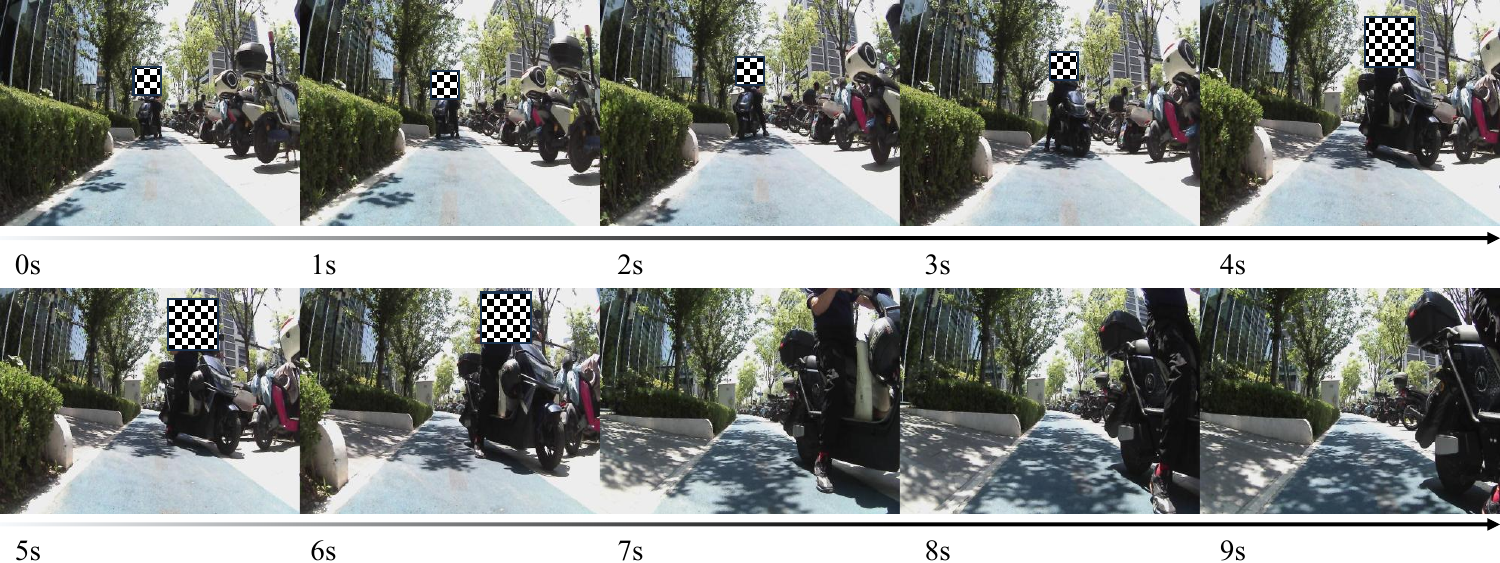}

    \vspace{-5pt} 
    \rule{\linewidth}{0.75pt}
    \vspace{-10pt} 
    
    \includegraphics[width=0.9\linewidth]{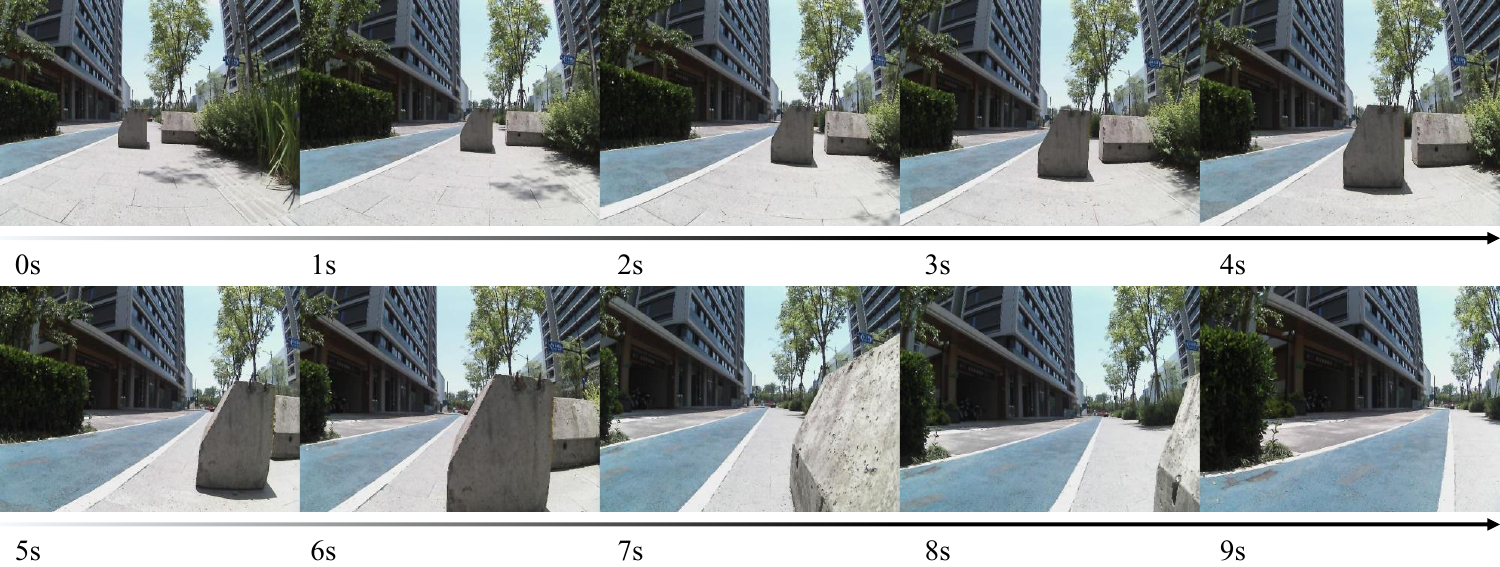}
    
    \caption{Qualitative results of real-world outdoor deployment under a \textbf{zero-shot} setting (without any task-specific training). Sensitive biometric information (e.g., human faces) has been anonymized.}
    \label{fig:outdoor_vis}
    \vspace{-10pt}
\end{figure}

\begin{figure}[t]
    \vspace{-20pt}
    \centering
    \includegraphics[width=0.9\linewidth]{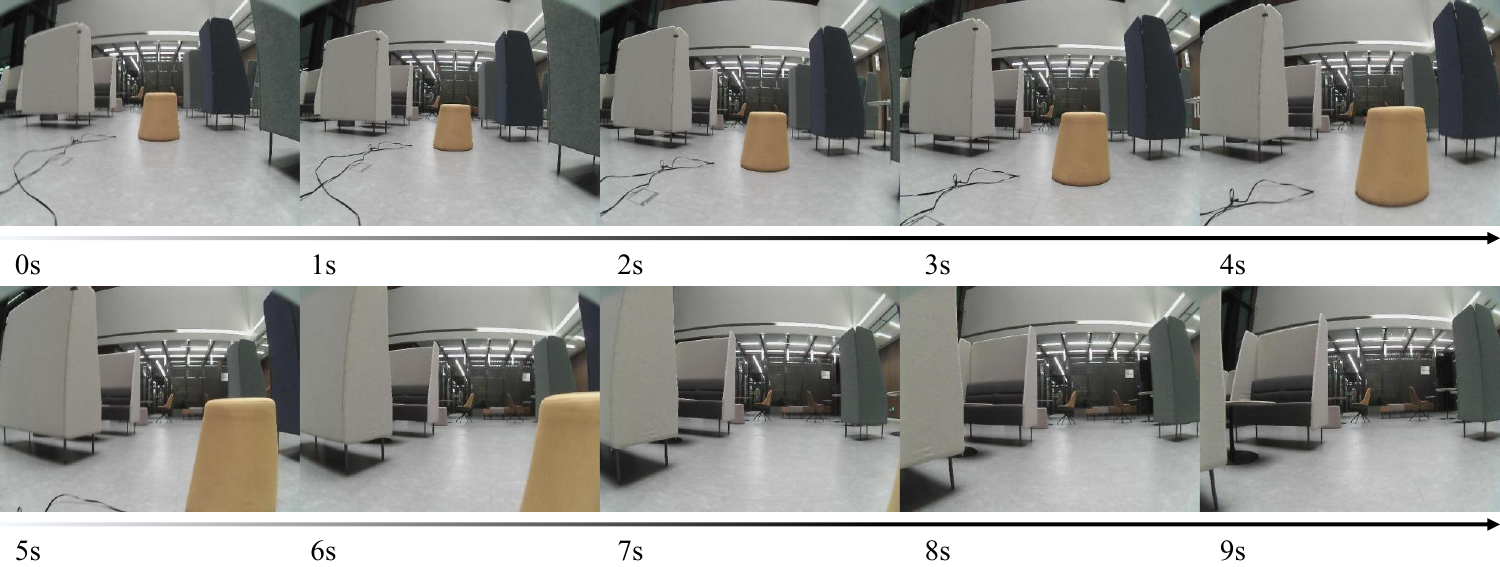}

    \vspace{-5pt} 
    \rule{\linewidth}{0.75pt}
    \vspace{-10pt} 
    
    \includegraphics[width=0.9\linewidth]{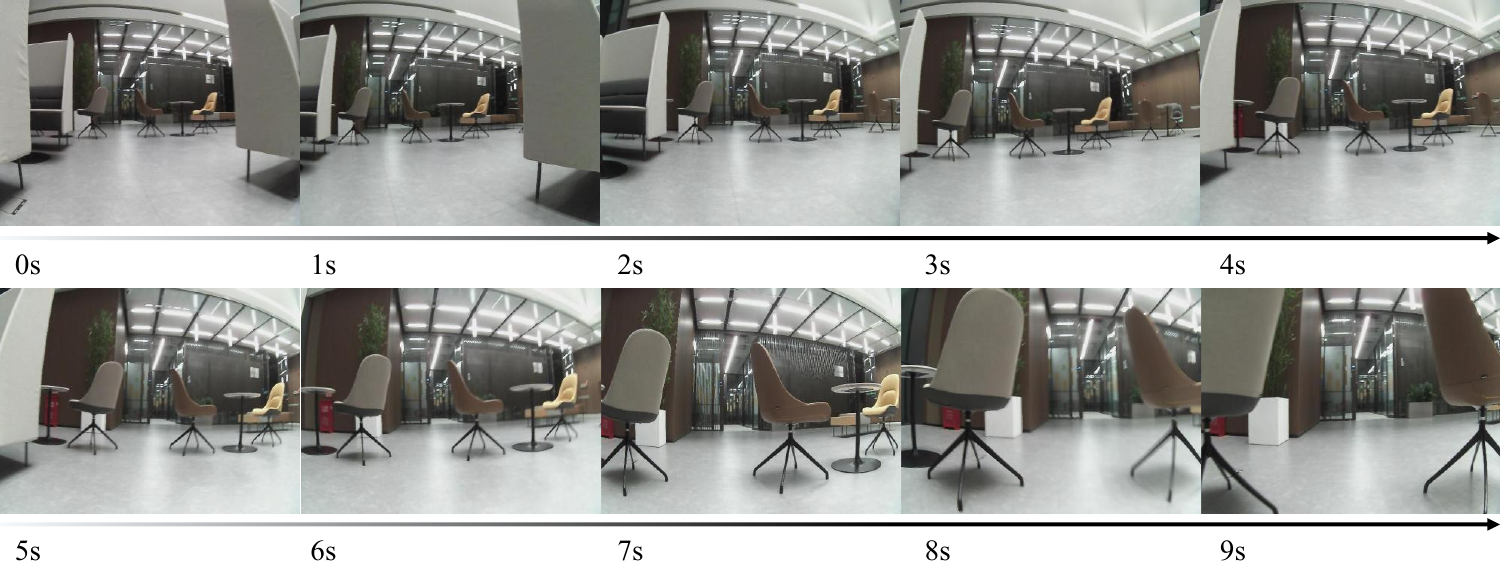}
    
    \caption{Qualitative results of real-world outdoor deployment under a \textbf{zero-shot} setting (without any task-specific training)}
    \label{fig:indoor_vis}
\end{figure}

We evaluate our method in closed-loop obstacle-avoidance tests under both indoor daytime and outdoor nighttime scenarios. The following four examples, as illustrated in Figure~\ref{fig:indoor_vis} and Figure~\ref{fig:outdoor_vis} the obstacle avoidance and generalization potential of our model. In the scenarios shown above, the quadruped robot is observed to plan appropriate paths, reflecting its capability to handle various environments. 
%

\section{Additional experiments results}

\begin{table*}[th]
\centering
\caption{\textbf{Performance comparison on NAVSIM-v1} \texttt{navtest} \textbf{Leaderboard}. 
* indicates training with reinforcement learning; -IL means imitation learning;
$\dag$ indicates training on the full navtrain split; 
$\ddagger$ indicates the use of the best-of-$N$ ($N=6$) strategy following~\cite{AutoVLA}.}
\label{tab:main_results_on_pdms_navtest}
\small
\resizebox{0.9\linewidth}{!}{
\begin{tabular}{l|c|cc|ccc|c}
\toprule
Method & Sensors & NC$\uparrow$ & DAC$\uparrow$ & EP$\uparrow$ & TTC$\uparrow$ & C$\uparrow$ & PDMS$\uparrow$ \\
\midrule

Human Agent & - & 100 & 100 & 87.5 & 100 & 99.9 & \cellcolor{gray!30} 94.8 \\

\midrule

\rowcolor[HTML]{EEE5F4}\hline
\multicolumn{8}{c}{\textit{Traditional E2E-based Planner}}\\

UniAD~\cite{uniad} & 6xC & 97.8 & 91.9 & 78.8 & 92.9 & \textbf{100.0} & \cellcolor{gray!30} 83.4 \\
TransFuser~\cite{transfuser} & 3xC+L & 97.7 & 92.8 & 79.2 & 92.8 & \textbf{100} & \cellcolor{gray!30} 84.0 \\
PARA-Drive~\cite{Para-drive} & 6xC & 97.9 & 92.4 & 79.3 & 93.0 & 99.8 & \cellcolor{gray!30} 84.0 \\
LAW~\cite{LAW} & 1xC & 96.4 & 95.4 & 81.7 & 88.7 & 99.9 & \cellcolor{gray!30} 84.6 \\
World4Drive~\cite{world4drive} & 3xC & 97.4 & 94.3 & 79.9 & 92.8 & \textbf{100.0} & \cellcolor{gray!30} 85.1 \\
DRAMA ~\cite{Drama} & 3xC+L & 98.0 & 93.1 & 80.1 & 94.8 & \textbf{100.0} & \cellcolor{gray!30} 85.5 \\
Hydra-MDP++~\cite{Hydra-mdp++} & 3xC+L & 97.6 & 96.0 & 80.4 & 93.1 & \textbf{100.0} & \cellcolor{gray!30} 86.6 \\
ARTEMIS~\cite{artemis} & 3xC+L & 98.3 & 95.1 & 81.4 & 94.3 & \textbf{100.0} & \cellcolor{gray!30} 87.0 \\
WorldRFT*~\cite{worldrft} & 3xC & 97.5 & 96.0 & 80.9 & 94.0 & \textbf{100.0} & \cellcolor{gray!30} 87.0 \\
DiffusionDrive~\cite{diffusiondrive} & 3xC+L & 98.2 & 96.2 & 82.2 & 94.7 & \textbf{100.0} & \cellcolor{gray!30} 88.1 \\
WorldDrive~\cite{worlddrive} & 1xC & 98.4 & 96.2 & 81.9 & 95.1 & \textbf{100} & \cellcolor{gray!30} 88.1 \\
WoTE~\cite{WoTE} & 3xC+L & 98.5 & 96.8 & 81.9 & 94.9 & 99.9 & \cellcolor{gray!30} 88.3 \\
SeerDrive~\cite{seerdrive} & 3/6xC+L & 98.4 & 97.0 & 83.2 & 94.9 & 99.9 & \cellcolor{gray!30} 88.9 \\
Drive-JEPA~\cite{drive-jepa} & 1xC & 98.7 & 96.2 & 82.9 & \textbf{100.0} & 95.5 & \cellcolor{gray!30} 89.0 \\


\rowcolor[HTML]{EEE5F4}\hline
\multicolumn{8}{c}{\textit{VLA-based Planner}}\\


AutoVLA-IL~\cite{AutoVLA} & 3xC & 96.9 & 92.4 & 75.8 & 88.1 & 99.1 & \cellcolor{gray!30} 80.5 \\
ReCogDrive-IL~\cite{recogdrive} & 1xC & 98.1 & 94.7 & 80.9 & 94.2 & \textbf{100.0} & \cellcolor{gray!30} 86.5 \\
SGDrive-IL~\cite{sgdrive} & 1xC & 98.6 & 95.1 & 81.2 & 95.4 & \textbf{100.0} & \cellcolor{gray!30} 87.4 \\
Vega~\cite{vega} & 1xC & \textbf{98.9} & 95.3 & 81.6 & 96.1 & \textbf{100.0} & \cellcolor{gray!30} 87.9 \\

\rowcolor[HTML]{EEE5F4}\hline
\multicolumn{8}{c}{\textit{WAM-based Planner}}\\

Epona~\cite{epona} & 1xC & 97.9 & 95.1 & 80.4 & 93.8 & 99.9 & \cellcolor{gray!30} 86.2 \\
ImagiDrive~\cite{li2025imagidrive}& 1xC &98.6 &96.2 &80.5 &94.5 &\textbf{100.0} &\cellcolor{gray!30} 87.4\\
PWM~$\dagger$~\cite{pwm} & 1xC & 98.6 & 95.9 & 81.8 & 95.4 & \textbf{100.0} & \cellcolor{gray!30} 88.1 \\
DriveVLA-W0~$\dag$~\cite{drivevla} & 1xC & 98.7 & 96.2 & 82.2 & 95.5 & \textbf{100.0} & \cellcolor{gray!30} 88.4 \\
UniWorldVLA~$\dagger$~\cite{uniworldval} & 1xC & 98.7 & 96.7 & 83.2 & 96.1 & \textbf{100.0} & \cellcolor{gray!30} 89.4 \\
DriveLAW~$\dagger$~\cite{drivelaw} & 1xC & 99.0 & 97.1 & 81.3 & 96.7 & \textbf{100.0} & \cellcolor{gray!30} 89.1 \\
\rowcolor{gray!20}\textbf{\ourMethod~(Ours)} & 1xC & 98.3 & 97.1 & 83.4 & 94.7 & \textbf{100.0} & \cellcolor{gray!30} 89.1 \\
\rowcolor{gray!20}\textbf{\ourMethod~$\ddagger$~(Ours)} & 1xC & 98.5 & \textbf{97.5} & \textbf{84.0} & 95.1 & \textbf{100.0} & \cellcolor{gray!30} \textbf{89.7} \\

\bottomrule
\end{tabular}}
\label{supp_pdms}
\end{table*}

\subsection{NAVSIM-v1 navtest Leaderboard.}
As shown in Table~\ref{supp_pdms}, under a fair comparison on NAVSIM-v1 NavTest, our method achieves state-of-the-art performance, reaching an overall PDMS of 89.7. It also significantly outperforms prior methods on both DAC and EP metrics. Notably, compared to previous WAM-based approaches, our method attains superior performance without using the full NavTrain dataset for training. This demonstrates that our paradigm of decoupling future image generation and action prediction effectively preserves the stability of the action space distribution, leading to smoother and more reliable driving behaviors.

\section{Ablation studies}

\subsection{Ablation of action expert capacity.}
We study the effect of the action expert capacity on planning performance. All variants share the same architecture as the video generation expert in terms of DiT depth, while varying the model size. As shown in Table~\ref{tab:ae}, increasing the capacity of the action expert consistently improves planning performance. The gains are particularly pronounced on challenging settings such as \texttt{navhard} S2, indicating that a larger action expert is beneficial for handling complex dynamics.

\subsection{Ablation of co-training between video generation and action prediction.}
While the action expert alone already achieves strong performance, jointly training with video generation further improves both PDMS and EPDMS by a clear margin. This validates that the proposed asymmetric attention mask enables the action expert to implicitly learn dynamic world evolution from the video generation branch during training, while remaining decoupled at inference time, thereby improving generalization.

\begin{figure}[t]
    \centering
    \begin{minipage}[t]{0.48\linewidth}
        \centering
        \captionof{table}{Ablation of action expert params.}
        \label{tab:ae}
        \resizebox{0.95\linewidth}{!}{
    \begin{tabular}{c|c|ccc}
    \toprule
    \multirow{2}{*}{\textbf{Params}} & \textbf{navtest} & \multicolumn{3}{c}{\textbf{navhard}} \\
    \cmidrule(lr){2-2} \cmidrule(lr){3-5}
    & EPDMS $\uparrow$ & S1 $\uparrow$ & S2 $\uparrow$ & EPDMS $\uparrow$ \\
    \midrule
    $\sim$ 0.21 B  & 88.2 & 76.0 & 40.7 & 31.2 \\
    $\sim$ 0.45 B  & 88.6 & 74.5 & 41.4 & 31.4 \\
    $\sim$ 1.04 B & \textbf{89.5} & \textbf{75.8} & \textbf{41.7} & \textbf{32.2} \\
    \bottomrule
    \end{tabular}
    }
    \end{minipage}
    \hfill 
    \begin{minipage}[t]{0.48\linewidth}
        \centering
        \captionof{table}{Ablation of training video generation expert.}
        \label{tab:training_paradigm}
        \resizebox{0.9\linewidth}{!}{
            \begin{tabular}{c|cc}
            \toprule
            Training Paradigm & PDMS & EPDMS \\
            \midrule
            w/o co-train  & 87.4 & 87.9 \\
            w/ co-train &  89.1& 89.5 \\
            \bottomrule
            \end{tabular}
        }
    \end{minipage}
\end{figure}

\begin{table}[t]
    \centering
    \caption{Performance and inference latency across different hardware and denoising steps.}
    \label{tab:denoise_step}
    \resizebox{0.7\linewidth}{!}{
        \begin{tabular}{c|ccc|cc}
        \toprule
        \multirow{2}{*}{\textbf{Steps}} & \multicolumn{2}{c|}{\textbf{navtest} $\uparrow$} & \textbf{navhard} $\uparrow$ & \multicolumn{2}{c}{\textbf{Latency (ms)} $\downarrow$} \\
        \cmidrule(lr){2-3} \cmidrule(lr){4-4} \cmidrule(lr){5-6}
        & PDMS & EPDMS & EPDMS & RTX 4090 & H200 \\
        \midrule
        1  & 87.0 & 87.2 & 30.4 & 110 & 100 \\
        2  & 88.9 & 89.2 & 31.2 & 147 & 140 \\
        5  & 89.0 & 89.4 & 31.4 & 280 & 240 \\
        10 & \textbf{89.1} & \textbf{89.5} & \textbf{32.2} & 480 & 430 \\
        \bottomrule
        \end{tabular}
    }
\end{table}

\section{Qualitative results}
We provide visualization results of the generation expert here, along with additional qualitative results on NAVSIM and CityWalker.

\begin{figure}[t]
    \centering
    \includegraphics[width=\linewidth]{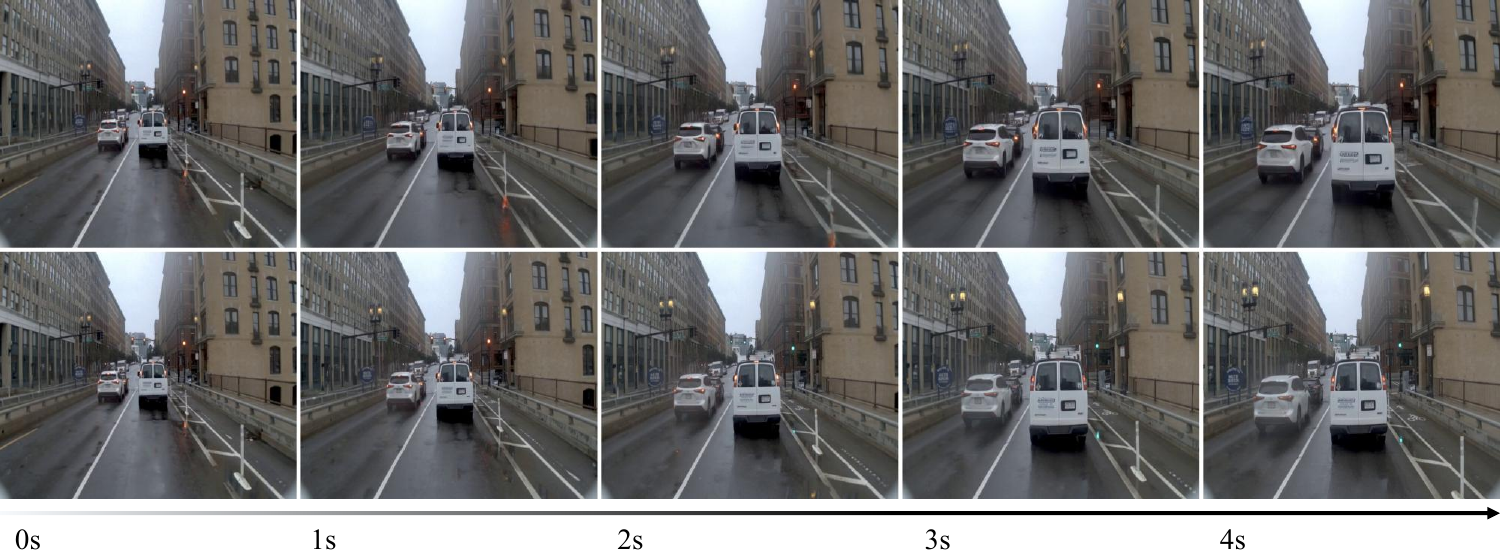}

    \rule{\linewidth}{0.75pt}
    \vspace{1pt} 
    
    \includegraphics[width=\linewidth]{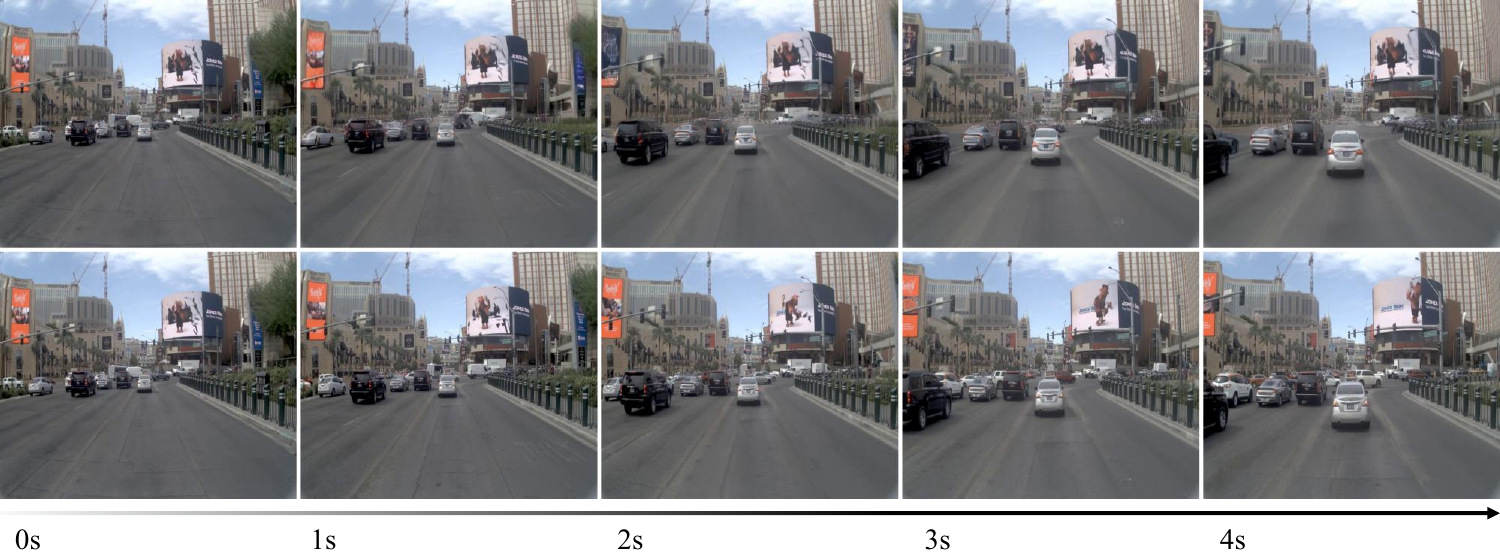}
    
    \caption{Qualitative comparison of video generation results. For each pair, the top row shows the frames generated by our model, and the bottom row shows the ground truth sequences.}
    \label{fig:vge_vis}
\end{figure}
\subsection{Qualitative of video generation expert}
As shown in Figure~\ref{fig:vge_vis}, we present visualization results of the generation expert in both static and dynamic scenarios. The model produces high-quality results in relatively static scenes. In contrast, under complex dynamic intersection scenarios, fine details of distant background vehicles may be lost. Nevertheless, this degradation does not noticeably affect short-term driving action prediction.

\begin{figure}[t]
    \centering
    \includegraphics[width=\linewidth]{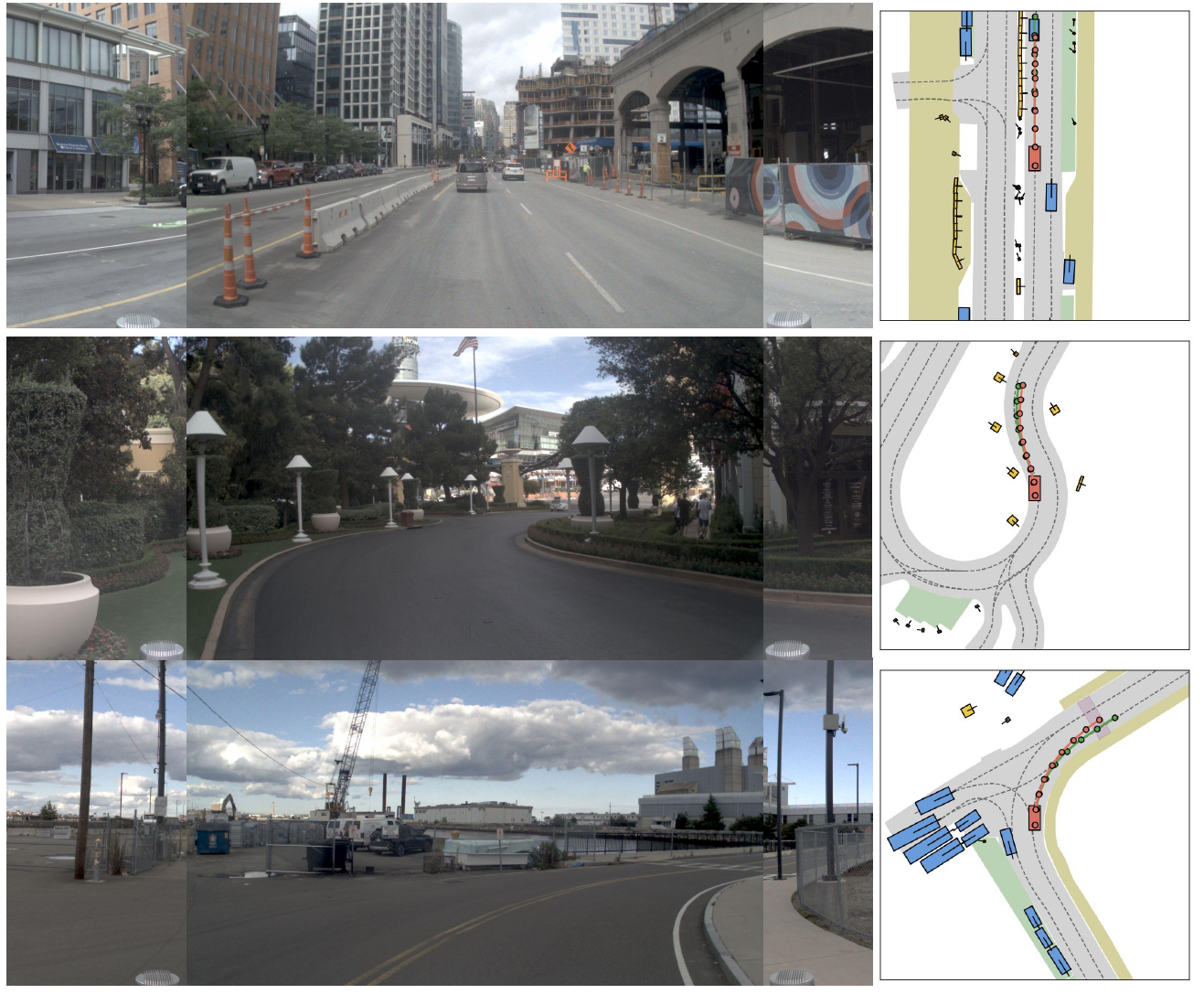}
    \caption{Qualitative results of \ourMethod in NAVSIM~\cite{NAVSIM}.}
    \label{fig:supp_navsim_vis}
\end{figure}
\begin{figure}[t]
    \centering
    \includegraphics[width=\linewidth]{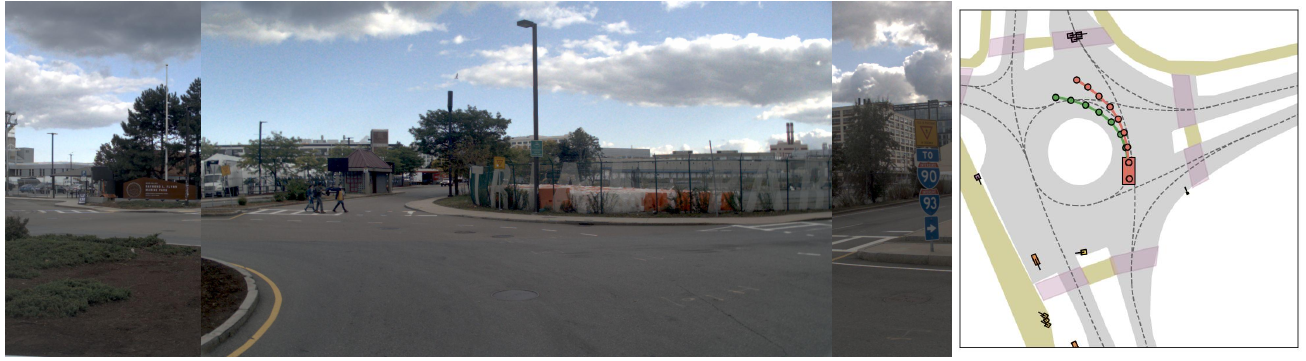}
    \caption{Qualitative failure results of \ourMethod in NAVSIM~\cite{NAVSIM}.}
    \label{fig:supp_navsim_failure}
\end{figure}
\subsubsection{More qualitative results and failure case on NAVSIM} 
Our method is capable of producing reasonable planning behaviors in scenarios such as lane following and turning, as illustrated in Figure~\ref{fig:supp_navsim_vis}. In open-world scenarios, our method may exhibit slight deviations during long-horizon planning due to the limitation of using only monocular-view inputs, as illustrated in Figure~\ref{fig:supp_navsim_failure}. We leave efficient multi-view world modeling for future work to further facilitate action planning.

\section{Experiment details and evaluation metric}

We adopt Wan2.2-5B as the VGE and the AE shares the same architecture with a reduced hidden dimension~($d_a = 1024$), resulting in a 1B-parameter branch and an overall model size of 6B.
For task setup, we use an action horizon of 8 for NAVSIMv2~(4 seconds with 0.5-second intervals), where each waypoint is represented as~$(x, y, \theta)$, and a horizon of 5 for CityWalker with~$(x, y)$. We use only the front-facing camera as input, and align video and action chunks with a 1:1 temporal ratio.

Both VGE and AE are trained under the same flow matching formulation. For NAVSIMv2, we use an input resolution of~$640 \times 768$ and train for 60 epochs with a batch size of 64; for CityWalker, we use~$384 \times 384$ and train for 30 epochs with the same batch size.
During inference, we use 10 denoising steps with classifier-free guidance~(CFG = 1.0). All experiments are conducted on 8 NVIDIA H200 GPUs (140 GB memory each).
We optimize all models using AdamW~($lr = 1 \times 10^{-4}$, weight decay 0.01) with a cosine annealing schedule, and apply mixed precision training with gradient clipping at 1.0.

We utilize two benchmarks in NAVSIMv2~\cite{NAVSIMv2} for end-to-end model evaluation, 
including \texttt{navhard} and \texttt{navtest}. \texttt{navhard} is the official two-stage evaluation benchmark, which contains 244 challenging real-world scenarios in the first stage and corresponding 4,164 synthetic scenarios generated by 3DGS in the second stage. \texttt{navtest} is a one-stage evaluation benchmark, containing a large number of 12,146 real-world scenarios. \texttt{navhard} focuses on assessing the model’s closed-loop performance in safety-critical situations, while \texttt{navtest} emphasizes generalization across diverse driving conditions. The two benchmarks share a rule-based planning metric, $\mathrm{EPDMS}$~\cite{Hydra-mdp++}, with several sub-metrics:
\begin{equation}
\mathrm{EPDMS} = 
\underbrace{\left(
\prod_{m \in \mathcal{M}_\text{pen}} S_m \right)
}_{\text{penalties}}
\cdot
\underbrace{\left(
\frac{ \sum_{m \in \mathcal{M}_\text{avg}} w_m  S_m }
     { \sum_{m \in \mathcal{M}_\text{avg}} w_m }\right)
}_{\text{weighted average}},
\label{eq:epdms}
\end{equation}
where $S_m$ is the sub-metric: penalty terms set $\mathcal{M}_\text{pen}$ includes No-at-fault Collisions (NC), Drivable Area Compliance (DAC), Driving Direction Compliance (DDC), and Traffic Light Compliance (TLC); weighted average terms set $\mathcal{M}_\text{avg}$ includes Time-to-Collision (TTC), Ego Progress (EP), Lane Keeping (LK), History Comfort (HC), and extended comfort (EC). Note that $\mathrm{EPDMS}$ in \texttt{navhard} further incorporates several modifications, two-stage aggregation, reactive traffic simulation, and the exclusion of penalties in cases where the human expert driver also fails.

We also provide one benchmarks in NAVSIMv1~\cite{NAVSIM} for end-to-end model evaluation on \texttt{navtest} with PDMS:
\begin{equation}
\text{PDMS} = \underbrace{\left( \prod_{m \in \{\text{NC}, \text{DAC}\}} \text{score}_m \right)}_{\text{penalties}} \times \underbrace{\left( \frac{\sum_{w \in \{\text{EP}, \text{TTC}, \text{C}\}} \text{weight}_w \times \text{score}_w}{\sum_{w \in \{\text{EP}, \text{TTC}, \text{C}\}} \text{weight}_w} \right)}_{\text{weighted average}}.
\end{equation}

\section{Limitation}
Despite its performance, our method has limitations. First, its reliability in extreme corner cases remains to be fully validated due to the inherent data-dependent nature of world modeling. Second, the framework relies heavily on the pre-trained Video Generation Expert; while video generation is bypassed during inference, its involvement in training remains computationally intensive. Additionally, the VAE's down-sampling ratio significantly impacts the balance between training efficiency and representation quality. Future work will focus on addressing these challenges to enhance robustness and training scalability.

\section{Broad Impact}
This research presents a dual-sided impact on the development of autonomous systems. On the positive side, by enabling agents to proactively reason about future environment evolution through world modeling, our framework significantly enhances the safety and decision-making efficiency of autonomous navigation, which may lead to a reduction in traffic accidents and energy consumption in smart cities. Conversely, the deployment of such technology entails potential risks: the model may exhibit unpredictable behavior in out-of-distribution (OOD) scenarios or extreme corner cases. Furthermore, its widespread adoption could disrupt the labor market for professional drivers and poses new challenges for legal frameworks regarding accountability in autonomous decision-making.


\end{document}